\documentclass[submission,copyright,creativecommons]{eptcs}

\providecommand{\event}{ARCADE 2019} % Name of the event you are submitting to
\providecommand{\titlerunning}{Using ConceptNet to Teach Common Sense to an Automated Theorem Prover}
\providecommand{\authorrunning}{Claudia~Schon, Sophie~Siebert \& Frieder~Stolzenburg}
\providecommand{\copyrightholders}{\authorrunning}
\providecommand{\publicationstatus}{}
\usepackage{underscore} % Only needed if you use pdflatex.

\usepackage{xcolor}
\usepackage{doc}
\usepackage{subcaption}
\usepackage[american]{babel}

\usepackage{tikz}
\usetikzlibrary{shapes.geometric} % required for the ellipse shape
\usetikzlibrary{arrows,backgrounds,calc,hobby,positioning}
\ExplSyntaxOn
\cs_if_exist:NF \prg_stepwise_function:nnnN { \cs_gset_eq:NN \prg_stepwise_function:nnnN \int_step_function:nnnN }
\cs_if_exist:NF \prg_stepwise_inline:nnnn { \cs_gset_eq:NN \prg_stepwise_inline:nnnn \int_step_inline:nnnn }
\ExplSyntaxOff
\tikzset{vertex style/.style={
    draw=#1,
    thick,
    text=black,
    ellipse,
    minimum width=0.5cm,
    minimum height=0.25cm,
    font=\footnotesize,
    outer sep=1pt,
  },
  text style/.style={
    sloped,
    text=black,
    font=\footnotesize,
    above
  }
}

\title{Using ConceptNet to Teach Common Sense\\to an Automated Theorem Prover%
  \thanks{The authors gratefully acknowledge the support of the German Research
	Foundation (DFG) under the grants SCHO~1789/1-1 and STO~421/8-1
	\emph{CoRg -- Cognitive Reasoning}.}}

\author{
  Claudia Schon
  \institute{Institute for Web Science and Technologies,
	University of Koblenz-Landau, 56070~Koblenz, Germany\\
	\email{schon@uni-koblenz.de}}
\bigskip
\and
  Sophie Siebert \qquad\qquad Frieder Stolzenburg
  \institute{Harz University of Applied Sciences,
  Automation and Computer Sciences Department,
  38855~Wernigerode, Germany\\
  \email{\{ssiebert,fstolzenburg\}@hs-harz.de}}
}

\begin{document}
\maketitle

\begin{abstract}
The CoRg system is a system to solve commonsense reasoning problems. The core of
the CoRg system is the automated theorem prover Hyper that is fed with large
amounts of background knowledge. This background knowledge plays a crucial role
in solving commonsense reasoning problems. In this paper we present different
ways to use knowledge graphs as background knowledge and discuss challenges that
arise.
\end{abstract}

\section{Introduction}
\label{sect:introduction}

\begin{figure}[t]
\begin{minipage}[b]{.5\textwidth}\em
	1: My body cast a shadow over the grass.\\
	What was the cause of this?\\
	A1: The sun was rising.\\
	A2: The grass was cut.
	\bigskip
	\caption{Problem~1 from the COPA benchmark set.}
	\label{fig:copa}
\end{minipage}
	\hfill
\begin{minipage}[b]{.45\textwidth}
    \includegraphics[scale=0.37]{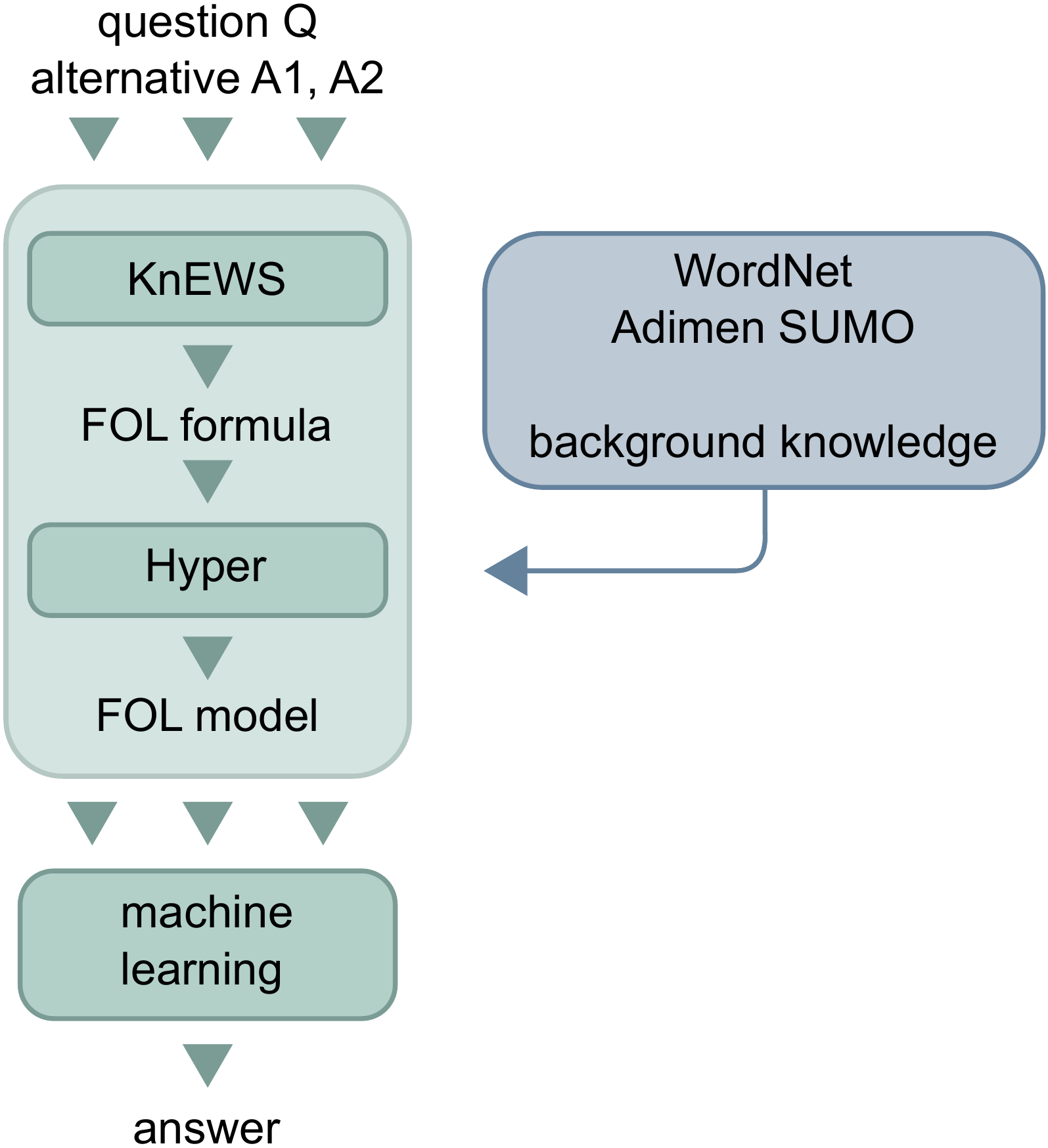}
    \centering
    \caption{The CoRg system.}
    \label{corg}
\end{minipage}
%\caption{COPA problem 1 together with some relevant knowledge from ConceptNet.}
%\label{fig:bla}
\end{figure}

In recent years, numerous benchmarks for commonsense reasoning have been
presented which cover different areas: the Choice of Plausible
Alternatives Challenge (COPA) \cite{roemmele2011choice} requires causal reasoning in
everyday situations, the Winograd Schema Challenge
\cite{DBLP:conf/aaaiss/Levesque11} addresses difficult cases of pronoun
disambiguation, the TriangleCOPA Challenge \cite{copa} focuses on human
relationships and emotions, and the Story Cloze Test with the ROCStories Corpora
\cite{storyclozetest} focuses on the ability to determine a plausible ending for
a given short story, to name just a few. In our system, we focus on the COPA
challenge where each problem consists of a problem
description (the premise), a question, and two answer candidates (called
alternatives). See Fig.~\ref{fig:copa} for an example. Most approaches tackling
these problems are based on machine learning or exploit statistical properties
of the natural language input (see e.g. \cite{semeval,RN+18}) and are therefore
unable to provide explanations for the decisions made.

%\begin{figure}
%\begin{tikzpicture}[node distance=2cm]
%\node[vertex style=black, ] (Rk) {sun};
% \node[vertex style=black, right=1.5cm of Rk] (light) {light}
% edge [<-,cyan!60!blue] node[text style]{Causes} (Rk); 
% \node[vertex style=black, right=1.5cm of light] (shadow) {shadow}
% edge [->,cyan!60!blue] node[text style]{AtLocation} (light); 
%\node[vertex style=black, below=0.3cm of light] (Skf) {ground}
% edge [<-,cyan!60!blue] node[text style]{AtLocation} (shadow);
%\node[vertex style=black, below=0.6cm of shadow] (Cf) {grass}
% edge [->,cyan!60!blue] node[text style]{AtLocation} (Skf);
%  \node[vertex style=black, right=1.5cm of shadow] (sunless) {sunless}
%   edge [->,cyan!60!blue] node[text style]{RelatedTo} (shadow); 
%  \node[vertex style=black, below=0.6cm of sunless] (sundial) {sundial}
%   edge [->,cyan!60!blue] node[text style]{RelatedTo} (shadow); 
%\end{tikzpicture}
%\caption{Some information in ConceptNet relevant for COPA problem 1.}
%\label{ex:conceptnet2}
%\end{figure}

In the CoRg project\footnote{\url{http://corg.hs-harz.de/}, accessed 2019-11-05}
(Cognitive Reasoning), we take
a different approach by using an automated theorem prover as a key component in
our system. Fig.~\ref{corg} gives an overview of the CoRg system: In the first
step, the problem description as well as the two 
alternatives
are each converted into first-order logic formulae (FOL) using KnEWS (Knowledge Extraction
With Semantics) \cite{knews}. KnEWS is a software that combines semantic
parsing, word sense disambiguation, and entity linking to produce a unified,
abstract representation of meaning. For example, the formula generated by KnEWS
for the first answer candidate from Fig.~\ref{fig:copa} (\emph{The sun was
rising.}) is:
\[
	\exists A (sun(A) \land \exists B (r1Actor(B,A) \land rise(B)))
\]

Each of these formulae is then passed to the automated theorem prover Hyper
\cite{cadesd} along with appropriate background knowledge. Currently, we use
Adimen-SUMO \cite{DBLP:journals/ijswis/AlvezLR12} and WordNet \cite{wordnet} as
background knowledge. Hyper computes a (possibly partial) model for each formula
together with the background knowledge. This (possibly partial) model contains
inferences performed by Hyper starting from the formula created for the natural
language sentences of the two possible answers. In the last step, the three
models created for the problem description and the answer candidates are further
processed by a machine learning component in order to decide which model points
more into the direction of the problem description. This includes a
preprocessing step and a deep learning neural network.

In the preprocessing step, the model has to be encoded such that the neural
network can process it. On the one hand, the logical facts in the model use
symbols corresponding to natural language words like $sun$ or $astronomicalBody$
which can be extracted. On the other hand, there is structural information in
the model given by the term structure of the logical facts. For instance, the
two facts $sun(sk0)$ and $is\_instance(sk0, \mathit{astronomicalBody})$ mean
that the sun is an astronomical body which is expressed via the Skolem constant
$sk0$. In our implementation, we currently only care about the symbols
corresponding to natural language words, i.e., $sun$ and
$\mathit{astronomicalBody}$. Currently, the structural information is dismissed.
Future work will address it as well.

The extracted natural language symbols form a sequence of words which are
transformed into their word embeddings. Word embeddings obtain semantic value by assigning
numerical values to words, thus making them comparable in a mathematical way.
For this, we use the pre-computed word embeddings from
ConceptNet~Numberbatch\footnote{\url{https://github.com/commonsense/conceptnet-numberbatch},
accessed 2019-06-12} with a dictionary of 400,000 words which are mapped into a
300 dimensional space. The resulting sequence of vectors are fed into a neural
network by building premise-answer pairs such that
each problem generates $n$ training examples with $n$ being the number of
alternatives to choose from. For the COPA benchmark set, it holds $n=2$.

\pagebreak

Eventually, our neural network has two inputs: one encodes the problem description while the other
encodes one of the answer candidates. In the core of the
neural network, the encoded information are merged together using an attentive
reader approach \cite{tan2015lstm} with bidirectional LSTMs (long short-term
memory) \cite{HS97}. The merging part consists of a dot-product between question
and answer to identify shared features between the texts. The following fully
connected dense layer generates an answer embedding with the context of the
question. This embedding is again merged with the encoded question using an add
operation. The output layer is a dense layer with 2 neurons and a softmax
activation such that it results in a vector $y^* = [y_1~y_2]$ with $y_1,y_2 \in
[0,1]$ and $y_1+y_2 = 1$ which can be interpreted as a likelihood of how well
the respective alternative fits to the given premise. The $n$ likelihoods of the
alternatives for one problem are compared and the highest one is chosen to be
the selected answer of our system \cite{SSS19b}. The inferences performed to
construct the model of the chosen answer can be used to provide an explanation
for the answer.

\section{Using Knowledge Graphs as Background Knowledge}
\label{sect:knowledgegraphs}

Besides ontologies like SUMO \cite{niles2001towards,Pease11}, Adimen SUMO \cite{DBLP:journals/ijswis/AlvezLR12}, Cyc \cite{lenat1995cyc} and Yago \cite{Suchanek:2008:YLO:1412759.1412998}, knowledge graphs constitute an important source of background
knowledge. The term \emph{knowledge graph} was coined by an initiative of the
same name by Google in 2012\footnote{\url{https://googleblog.blogspot.com/2012/05/introducing-knowledge-graph-things-not.html}, accessed 2019-11-05} and is now widely used as a generalized term for
interlinked descriptions of entities.
Compared to ontologies, these knowledge graphs usually contain mainly factual
knowledge as triples of the form $(s,p,o)$ (subject -- predicate -- object).
However, they contain very large amounts of this factual knowledge. Examples for
knowledge graphs are BabelNet \cite{NavigliPonzetto:12aij} and ConceptNet
\cite{DBLP:conf/aaai/SpeerCH17}. BabelNet was automatically created by linking
Wikipedia to WordNet, a lexicon of the English language.
ConceptNet is a freely-available semantic network designed to help computers understand the meanings of words that people use. Large parts of ConceptNet were created by humans which is why ConceptNet contains interesting commonsense knowledge.
One example for knowledge represented in ConceptNet that is difficult to find in other sources is the following triple: 
\[
(\mathit{snore},\mathit{HasSubevent},\mathit{annoy\ your\ spouse})
\]
Since this kind of knowledge is hardly present in other knowledge bases, we would like to use ConceptNet as a source for background knowledge in the CoRg project. 

If knowledge represented in a knowledge graph like ConceptNet is supposed to be
used by a first-order logic theorem prover, the triples have to be translated
into first-order logic. The easiest way to do that would be to translate $p$
into a predicate name and both $s$ and $o$ into constants leading to $p(s,o)$.
Since this is only factual knowledge, it is only of limited use for the
commonsense reasoning task under consideration. Due to the fixed set of
predicates used in ConceptNet it is possible to create translations of
ConceptNet triples to first-order logic formulae depending on the predicate used
in the triple. Another way to translate a triple $(s,p,o)$ given in ConceptNet
into the first-order logic formula would be:
\[
\forall x \Bigr(s(x) \rightarrow \bigr(\exists y (p(x,y) \land o(y))\bigl)\Bigl)
\]

\pagebreak

Fig.~\ref{ex:conceptnet} shows some information in ConceptNet relevant for COPA
problem 1 (see Fig.~\ref{fig:copa}). The triples given implicitly in
Fig.~\ref{ex:conceptnet} would be translated into:\vspace*{-0.3cm}

\begin{figure}[h!]\hspace*{-0.5cm}
\begin{minipage}{.3\textwidth}\noindent
	\begin{eqnarray*}
	&\forall x \Bigr(sun(x) \rightarrow \bigr(\exists y (causes(x,y) \land light(y))\bigl)\Bigl)\\
	&\forall x \Bigr(shadow(x) \rightarrow \bigr(\exists y (atlocation(x,y) \land light(y))\bigl)\Bigl)\\
	&\forall x \Bigr(shadow(x) \rightarrow \bigr(\exists y (atlocation(x,y) \land ground(y))\bigl)\Bigl)\\
	&\forall x \Bigr(grass(x) \rightarrow \bigr(\exists y (atlocation(x,y) \land ground(y))\bigl)\Bigl)
	\end{eqnarray*}
\end{minipage}
	\hfill
\begin{minipage}{.45\textwidth}
	\begin{tikzpicture}[node distance=2cm]
	\node[vertex style=black, ] (Rk) {sun};
	 \node[vertex style=black, right=1.5cm of Rk] (light) {light}
	 edge [<-,cyan!60!blue] node[text style]{Causes} (Rk); 
	 \node[vertex style=black, right=1.5cm of light] (shadow) {shadow}
	 edge [->,cyan!60!blue] node[text style]{AtLocation} (light); 
	\node[vertex style=black, below=0.3cm of light] (Skf) {ground}
	 edge [<-,cyan!60!blue] node[text style]{AtLocation} (shadow);
	\node[vertex style=black, below=0.6cm of shadow] (Cf) {grass}
	 edge [->,cyan!60!blue] node[text style]{AtLocation} (Skf);
	\end{tikzpicture}
	\caption{Some information in ConceptNet relevant for COPA problem 1.}
	\label{ex:conceptnet}
\end{minipage}
\end{figure}
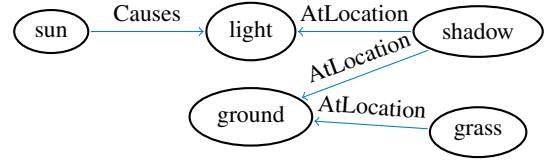

At first glance, this translation looks quite promising. On closer inspection, however, one realizes that starting from a fact $sun(a)$, it is not possible for a constant $a$ to derive something using the predicate $shadow$.
The problem is that the direction of the edges in ConceptNet affects the generated FOL formulas. One possible solution would be to generate two formulas for each edge in ConceptNet. For example, for the edge from \emph{light} to \emph{shadow} we could generate the additional formula:
\[
\forall x \Bigr(light(x) \rightarrow \bigr(\exists y (\mathit{maysituate}(x,y) \land shadow(y))\bigl)\Bigl)
\]
Another problem is the large amount of information available in ConceptNet. The
information in Fig.~\ref{ex:conceptnet} presents a manually selected part of
ConceptNet relevant for problem~1 of the COPA challenge. In total, node
$sun$ has 637 incoming and 1,000 outgoing edges and node $shadow$ 399 incoming
and 626 outgoing edges in ConceptNet. It is not trivial to select from this
variety of edges those that are relevant to the problem under consideration.

To solve this problem, we plan three things: 
\begin{itemize}
  \item We will only consider a subset of the relations used in ConceptNet.
	ConceptNet uses a fixed set of around 40 relations in its triples.
	Examples for these relations are general relations like $\mathit{IsA}$
	and $\mathit{PartOf}$ as well as more specific relations like
	$\mathit{CapableOf}$ and $\mathit{Desires}$. All relations can be
	negated by prefixing them with the word \emph{Not}. Many of these
	relations are not interesting for our purpose. %For example
	The relation \emph{ExternalURL} can be used to point to an URL outside
	of ConceptNet where further linked data about a certain node can be
	found. Furthermore, there are relations providing information relevant
	for languages other than English. This is why we plan to manually
	selected set of relations that are interesting for the COPA problems.
\item Despite the restriction to a subset of the relations used in ConceptNet, the formula set generated from ConceptNet is likely to be too large. Therefore, we will try to select suitable formulas from this formula set. Here we will experiment with SInE \cite{Hoder:2011uq} and Similarity SInE \cite{CADE27}.
  \item Starting from the manually selected relations from ConceptNet, we only
	consider the triples whose third component is similar to the words in
	the COPA problem currently under consideration. To measure the
	similarity we are planning to use word embeddings like ConceptNet
	Numberbatch. This would result in not using the triple $(sun,IsA,star)$
	for COPA problem~1 (cf. Fig.~\ref{fig:copa}), since none of the words in COPA problem one is very
	similar to the word star. In contrast to that, $(sun,Causes,light)$
	would be used.
\end{itemize}

\section{Conclusion\,/\,Future Work}
\label{sect:conclusion}

In this paper, we introduced the CoRg system which is able to tackle commonsense
reasoning problems by combining a first-order logic theorem prover, background
knowledge bases, and machine learning. We discussed how to integrate knowledge
represented in a knowledge graph into the CoRg system such that the theorem
prover is able to use this knowledge. In future work, we plan to investigate how
to deal with the vast amount of triples in knowledge graphs. In addition
to that, we would like to integrate other knowledge graphs like e.g.
BabelNet \cite{NavigliPonzetto:12aij} into our system for commonsense reasoning.

\label{sect:bib}
\bibliographystyle{eptcs}
\bibliography{lit}

\end{document}